\documentclass[letterpaper,twocolumn,10pt]{article}
\usepackage{zhanggroup}

\usepackage{tikz}
\usepackage{graphicx}
\graphicspath{{figure/}}
\usepackage{multirow}
\usepackage{array}
\usepackage{seqsplit}
\usepackage[most]{tcolorbox}
\usepackage{xurl}
\usepackage{amsfonts}

\tcbset{cvprappendixlisting/.style={
  width=\dimexpr\linewidth-10pt\relax,
  center,
  shrink break goal=4pt
}}

\newcommand{\FakeVLM}{\textsc{FakeVLM}}
\newcommand{\SIDA}{\textsc{SIDA-13B}}

\newcommand{\CommFor}{\textsc{CommForensics}}
\newcommand{\AIDE}{\textsc{AIDE}}
\newcommand{\FakeShield}{\textsc{FakeShield}}

\newcommand{\Luna}{\textsc{GPT-5.6-luna}}
\DeclareRobustCommand{\code}[1]{\texttt{\protect\seqsplit{#1}}}
\DeclareRobustCommand{\toolname}[1]{\texttt{\protect\seqsplit{#1}}}

\title{WildFakeToolBench: Benchmarking AI-Generated Image Detectors with Web-Sourced Image Processing}
\author{
Mingjie Li\textsuperscript{1}\ \ \
Yukun Jiang\textsuperscript{1}\ \ \
Yicong Tan\textsuperscript{1}\ \ \
Yang Zhang\textsuperscript{1}
\\
\\
\textsuperscript{1}\textit{CISPA Helmholtz Center for Information Security}
}
\begin{document}
\maketitle
\begin{abstract}
As AI generated images are getting popular these days, discussions on such generations are widely existed in current public forums, software repositories, and photography tutorials. 
Moreover, with the help of LLM-assisted code translation, malicious actors can adopt such information to evade detectors without specialized knowledge or local GPU execution of the resulting tools. 
To evaluate such risks, we introduce \textsc{WildFakeToolBench}, a benchmark suite of $156$ executable image-processing tools derived from $220$ public sources across $142$ domains. 
Each tool retains its source evidence and passes syntax, execution, and output-distortion checks without using feedback from the evaluated detectors. 
On $105$ generated images from five generators, we evaluate six state-of-the-art detectors with images processed by the tools in our collected bench to assess their effectiveness against such in-the-wild risks.
The strongest individual tools achieve attack success rates of $25.25\%$--$100\%$ across detectors, exceeding the strongest of six selected image-processing baselines for each detector. 
Evaluating subsets of an eight-tool pool yields worst-case coverage above $80\%$ for every detector. 
Effectiveness is sparse and detector-specific: high coverage across candidates does not imply reliable one-shot evasion. 
These findings not only highlight the potential of such web-sourced risks, but also show the potential on establishing AI-image detector robustness evaluation framework based on those in-the-wild information. Code and benchmark can be found in \url{https://github.com/TrustAIRLab/WildFakeToolBench}.
\end{abstract}
\section{Introduction}
As AI-generated images become increasingly accessible, more users are discussing them online. However, these images can also be used for malicious purposes, such as spreading misinformation. Although several detectors have been proposed to identify statistical, frequency-domain, or visual-artifact cues, prior work has shown that their robustness against various attacks remains limited~\cite{ERCBSPCBFCB25,HD19,RBS20}. Moreover, malicious actors can draw on publicly accessible web knowledge about photographic appearance, camera artifacts, and detector evasion. With the help of current LLMs, they can translate this knowledge into image-processing operations that make generated images evade detectors. Because these operations can be applied by non-experts with image-processing knowledge using only a CPU, this risk should not be overlooked.

Such knowledge is distributed across forums, software repositories, image-tool sites, and photography tutorials. Some sources explicitly discuss evasion, while others describe ordinary editing or simulated capture effects. Although these sources often provide informal descriptions rather than directly applicable code or software, attackers can use LLMs to translate them into runnable code, reducing the implementation expertise required. Most resulting transformations can also be performed with standard image-processing libraries on CPUs. The relevant question is therefore whether accessible public experience can supply effective robustness tests for current detectors, even without detector-specific attack optimization.

\begin{figure*}[t]
\centering
\includegraphics[width=1.0\textwidth]{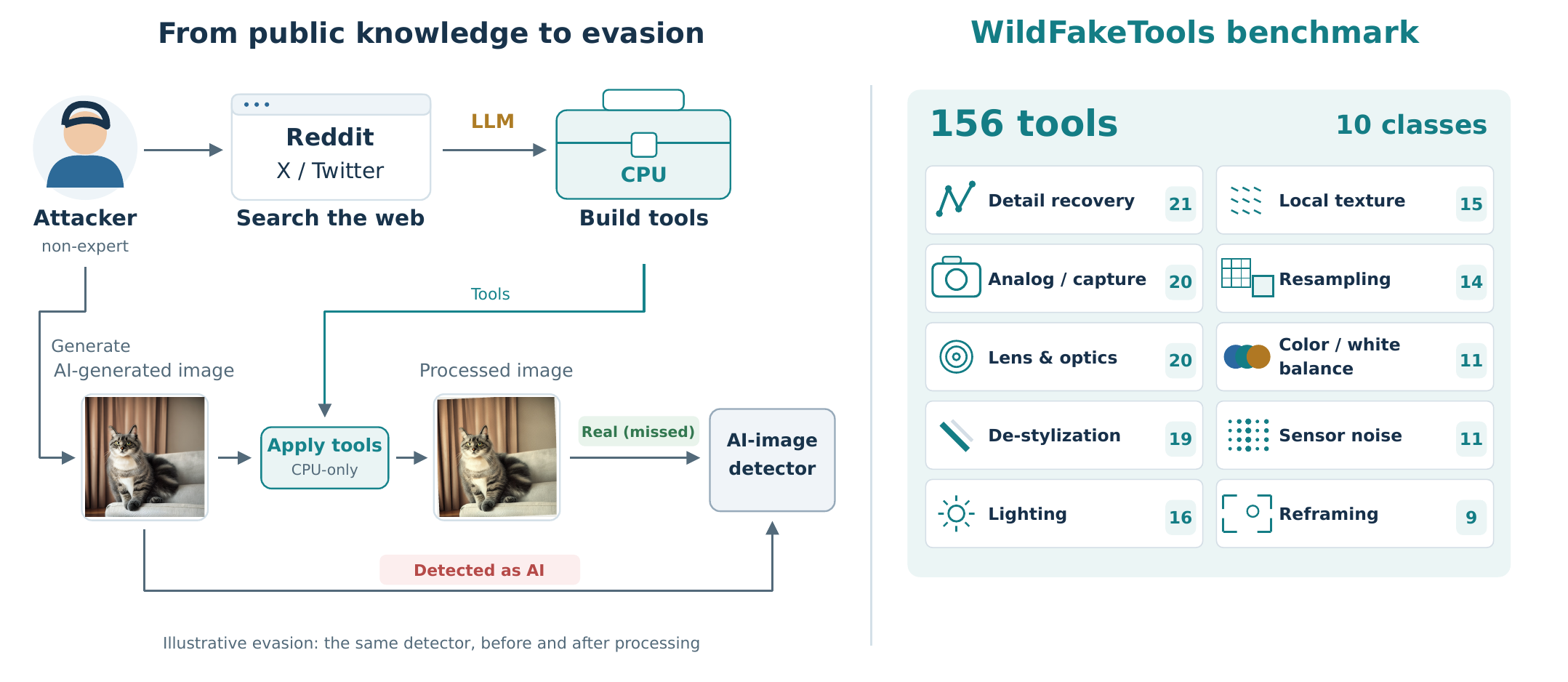}
\caption{\textsc{WildFakeToolBench} overview. Threat Model (Left): a non-expert user searches public web experience and uses LLM-assisted translation to obtain CPU-compatible image-processing tools for detector evasion. WildFakeToolBench (Right): we organize the recovered methods into a benchmark of $156$ tools across ten processing classes. The paired images illustrate the software \toolname{print\_rescan} transformation.}
\label{fig:benchmark_overview}
\end{figure*}

We introduce \textsc{WildFakeToolBench}, a suite of $156$ executable tools with source provenance and a common evaluation protocol. The tools are generated by our WildFake agent framework, which performs text retrieval, method extraction, deduplication, and code generation. They span ten categories, including artifact repair, color adjustment, resampling, and simulated capture effects. To reflect attackers with limited resources, the image-processing operations in our benchmark require no local GPU. Neither tool synthesis nor image-specific parameter proposals use detector feedback. Our contribution is the resulting benchmark resource and its assessment of an accessible processing space, rather than a new detector-specific optimization algorithm.

To measure the robustness of current state-of-the-art AI detectors in our settings, we evaluate six state-of-the-art AI detectors after processing our collected tools on an image set collected across five generators and seven content categories. 
For each tool and image, we process the tool with five parameter candidates generated by GPT-4o-mini and retain outputs with SSIM at least $0.75$. 
Based on the evaluation, we find that every evaluated detector is vulnerable to at least one tool, and the strongest individual tool exceeds the strongest of six selected baselines for each detector.
Although most tools cannot effectively affect the prediction of the state-of-the-art detectors, we still find many useful tools for current detectors. For example, the maximum single-tool attack success rate (ASR) for \FakeVLM{} is $45.54\%$, demonstrating substantial weaknesses for current detectors.

Besides single-tools, malicious actors may also collect several attack tools, combine them for effective attacks. 
Therefore, we also evaluate the coverage obtained by trying subsets of several available tools. 
It exposes broader weaknesses, but measures a worst-case candidate portfolio rather than one-shot evasion. 
Since the combination of all tools are too many, we select a candidate pool from our single-tool evaluation to our coverage evaluation. 
Besides the measurement on the risks, our in-the-wild analysis also traces how public sources contribute executable methods and how tool categories and source types appear among effective candidates. Our contributions are threefold:
\begin{itemize}
\item We introduce \textsc{WildFakeToolBench}, a benchmark of $156$ executable, provenance-linked image-processing tools derived from public web knowledge, with CPU-compatible implementations and shared validity checks.
\item We evaluate six AI-image detectors under single-tool and multiple-candidate protocols, revealing substantial but detector-specific weaknesses beyond six selected image-processing baselines.
\item We analyze the conversion of public knowledge into tools, the processing strategies it covers, and the sources represented among effective tools, providing a basis for reproducible robustness testing.
\end{itemize}

\section{Related Work}
\subsection{AI Image Generators}

AI image generation has progressed through several modeling paradigms, making
high-quality synthesis increasingly accessible. Early generative adversarial
networks~\cite{GPMXWOCB14,MO14,WLZTKC18} often produced lower-fidelity images. Therefore, diffusion~\cite{HJA20,SME21} and latent-diffusion models~\cite{RBLEO22} subsequently became dominant,
particularly for high-fidelity text-to-image synthesis~\cite{PELBDMPR23,sd21}, and attracted
substantial public attention. 
However, diffusion models can still struggle to follow complex user prompts, limiting their utility for some practical tasks.
Newer generation systems, such as GPT-Image and Nano Banana~\cite{GPT-Image-1, NanoBanana2, GPT-Image-1.5}, improve prompt understanding with their internal language models~\cite{BMRSKDNSSAAHKHCRZWWHCSLGCCBMRSA20,O23} while producing high-quality images. 
This evolution is relevant to detection because different generator families may leave different color, spectral, or processing artifacts. 
We therefore include five popular generators in our evaluation.

\subsection{AI-Generated Image Detectors}
AI-generated images can be used to spread rumors and misinformation online,
potentially causing substantial social harm. Researchers have therefore
developed detectors to distinguish AI-generated images from real ones~\cite{HPBSSZZ24}.
Traditional neural-classifier-based detectors are commonly trained on large
datasets of generated images~\cite{SLYZ22,PO25} or on specific artifacts
associated with image generation~\cite{YLCHJHX25,YDF19,FESFKH20}. Some methods leverage the
generator models themselves for zero-shot fake-image
classification~\cite{STLBZ24}.

More recently, the strong visual understanding capabilities of vision-language
models have motivated VLM-based detectors. \FakeVLM{}~\cite{WYFKWCWWHL25} is
fine-tuned to reason about visual artifacts associated with generated images.
\SIDA{}~\cite{HHLHZPWHC25} and \FakeShield{}~\cite{XZLTHZ25} further combine
classification with localization or segmentation of manipulated regions, while ViGText~\cite{ANKKP26} also adopt graph neural networks for analysis. 
In our evaluation, we assess the robustness of these detectors against tools
derived from in-the-wild knowledge.

\section{WildFakeToolBench Benchmark}
\label{sec:collection}

\subsection{Threat Model}
As AI models are easily accessible for common people, we now consider a non-professional, in-the-wild malicious actor who seeks to make an AI-generated image evade detection by having it classified as real. 
The actor can search publicly available online knowledge and use a general-purpose LLM to translate relevant advice into image-processing tools, but is not assumed to have specialized image-forensics expertise or local GPU resources. 
The actor can submit images to a black-box detector and observe its binary decision (real or AI-generated). 
The detector's identity and all other information—including its architecture, parameters, training data, confidence scores, and decision threshold—are unknown to the actor. Thus, the threat model permits only label-level feedback, not white-box access or detector-specific knowledge.

\subsection{Benchmark Collection Framework}

To collect web-based tools can be utilized by malicious actors, we build a agent framework called \textsc{WildFake} by simulating the actors' operation on collecting knowledge and converting them to applicable tools.
The framework contains three stages as follow,
\begin{itemize}
  \item \textbf{Retrieval and Knowledge Processing:} We first adopt Tavily to collect knowledge about AI-generated images from public sources. Then we adopt GPT-4o-mini to preprocess the collected texts, removing the low quality terms like short and out-of-scope texts. And we finally formulate them in a constant template for provenance.
  \item \textbf{Method Extraction and Deduplication:} After collected the information, we adopt GPT-4o-mini to extract actionable methods and other information related to the method for further analysis, like quoted evidence. Then we also adopt the model to identify the repeated techniques and merge them to preserve the unique tools for our benchmark.
  \item \textbf{Tool Synthesis and Validation:} After extracted the methods, we adopt GPT-4o-mini to convert the retained method into a parameterized Python image-processing tool. Since our threat models targets on the non-professional attackers, our generated image-processor are all applicable for CPU tools as such attackers usually do not have sufficient GPU resources for diffusion or other techniques. We also check the tools by evaluating their interface validity, execution, non-identity output, and distortion. Failures can be revised using software diagnostics without detector feedback.
\end{itemize}

The details of our agent framework are listed in Appendix~\ref{app:construction}.

\subsection{The Obtained WildFakeToolBench}
As shown in the right part of Figure~\ref{fig:benchmark_overview}, our obtained \textsc{WildFakeToolBench} contains $156$ validated tools derived from $220$ public sources across $142$ domains. These synthesized image-processing operations use standard libraries and require no local GPU.
The tools retain source provenance and executable implementations, and are organized into ten categories spanning operations such as color adjustment, resampling, and simulated capture effects, details are described in Appendix~\ref{app:categories}.
Their goals fall into broader patterns: reducing recognizable synthetic appearance through detail repair, de-stylization, and texture changes; adjusting lighting and color; adding simulated capture cues through analog effects, lens effects, noise, and reframing; and reproducing resizing or re-encoding associated with image sharing. 

\paragraph{Example: Translating Capturing Techniques.}
A Reddit discussion describes printing an image and scanning it back. We list its source, idea and implementation as below. We note it is a software simulation, not a physical print-and-scan experiment. Its $25.25\%$ ASR against \SIDA{} illustrates how public advice can yield a runnable test without detector-specific optimization. 
\begin{tcolorbox}[
  breakable,
  colback=white,
  colframe=gray!60,
  boxrule=0.5pt,
  arc=1pt,
  left=4pt,
  right=4pt,
  top=3pt,
  bottom=3pt,
  title={Example: \toolname{print\_rescan}},
  fonttitle=\bfseries
]
\textbf{Source idea:} Print the image and scan it back in. The synthesized tool approximates this operation in software.

\textbf{Source:} Reddit ``How to prevent AI image
        detection?''\footnote{https://www.reddit.com/r/StableDiffusion/comments/13eqp8v/}.

\textbf{Software implementation.} The tool uses paper-fibre noise with low-frequency brightness variation to simulate printed paper, gamma and color
adjustments to reproduce printing and scanning shifts. It also adopts a small rotation to mimic the scan misalignment, and Gaussian blur to model recapture softness.
\end{tcolorbox}

\subsection{Settings of our Benchmark Evaluation}
\label{sec:eval}

\paragraph{Detectors and Baselines}
\label{sec:eval-detectors}
We evaluate four VLM-based systems, \FakeVLM{}~\cite{WYFKWCWWHL25}, \SIDA{}~\cite{HHLHZPWHC25}, \FakeShield{}~\cite{XZLTHZ25}, and \Luna{}~\cite{gpt5_luna}, and two neural classifiers, \CommFor{}~\cite{PO25} and \AIDE{}~\cite{YLCHJHX25}. We use \FakeShield{}'s DTE-FDM component for classification and prompt \Luna{} directly. Additional detector descriptions and the prompt appear in Appendices~\ref{app:detectors} and~\ref{sec_gpt}.

We compare against six selected image-processing baselines: Gaussian/salt-and-pepper noise, spectral suppression, defocus blur, Poisson--Gaussian noise, simulated camera processing, and Reinhard color transfer~\cite{HD19,WRAQ22,FTKE08,BMXCSB19,RAGS01}. All use the same candidate and distortion-filtering protocol. This comparison measures additional stress-testing value relative to these implementations, rather than superiority over all academic attacks. Details appear in Appendix~\ref{app:baselines}.

\paragraph{Evaluated Images}

Our image set for evaluation contains $105$ generated images: $21$ shared prompts submitted to Stable Diffusion 2.1~\cite{sd21}, Qwen-Image-2.0~\cite{T26}, Midjourney~\cite{Midjourney}, GPT Image 2~\cite{GPT-Image-2}, and Nano Banana 2~\cite{NanoBanana2}. Following prior work~\cite{WYFKWCWWHL25}, prompts span animal, object, document, human, scene, satellite, and face-portrait categories, with three prompts per category. GPT-4o-mini proposes prompts, which are manually reviewed. For each prompt--generator pair, we generate three candidates and manually select the highest-quality output. This controlled, curated set supports paired processing comparisons but does not represent the full distribution of generated images.

\paragraph{Single-Tool Protocol}
\label{sec:eval-protocol}
For each tool and image, GPT-4o-mini proposes five parameter settings spanning processing strengths, before any detector query. We discard outputs with SSIM~\cite{WBSS04} below $0.75$ relative to the input. This threshold controls measured structural distortion, without guaranteeing naturalness or semantic preservation. If all candidates fail the filter, the attack fails on that image.

Let $\mathcal{B}_d$ be the images initially classified as fake by detector $d$. A tool succeeds on $x\in\mathcal{B}_d$ if any retained candidate is classified as real. ASR is the fraction of $\mathcal{B}_d$ successfully attacked. Table~\ref{tab:single_tool_asr} reports the detector-specific denominators, so baseline detection errors are not credited as attacks. For example, \FakeVLM{} detects $101$ of $105$ original images (over $96\%$), with four errors. Different denominators also limit direct ASR comparisons between detectors.

\paragraph{Multiple-Tool Coverage}
For a set $\mathcal{T}$ of $k$ available tools, we evaluate every non-empty subset $c\subseteq\mathcal{T}$. For each image and subset, GPT-4o-mini proposes one execution order and five joint parameter settings, without detector feedback. We do not enumerate alternative orders. Let $\mathcal{Y}(x,c)$ be the outputs passing the same SSIM filter. Stacked ASR measures candidate coverage.
Let $E_d(x,\mathcal{T})$ be the event that some non-empty subset $c\subseteq\mathcal{T}$ yields $y\in\mathcal{Y}(x,c)$ with $d(y)=\textsc{Real}$. Then
\begin{equation}
\mathrm{ASR}^{\mathcal{T}}_d =
\frac{1}{|\mathcal{B}_d|}\sum_{x\in\mathcal{B}_d}
\mathbb{1}\!\left[E_d(x,\mathcal{T})\right].
\label{eqn:stack_asr}
\end{equation}
This is an oracle, worst-case coverage measure over evaluated candidates: different images may succeed with different subsets. There are up to $5(2^k-1)$ candidates per image before filtering, or $1{,}275$ at $k=8$. It does not measure one-shot success, require all $k$ tools to be applied, or establish a practical tool-selection policy.

Exhaustive search over $156$ tools is infeasible. We form a fixed eight-tool pool from the top two single tools for \FakeVLM{},\SIDA{},\FakeShield{}, \CommFor{} and \AIDE{}, using their evaluation ASRs. Results therefore describe a performance-selected pool on this image set. Tool construction and parameter proposals remain independent of detector feedback; pool selection and identification of the best successful candidate do not.

\begin{figure}[!t]
\centering
\includegraphics[width=0.45\textwidth]{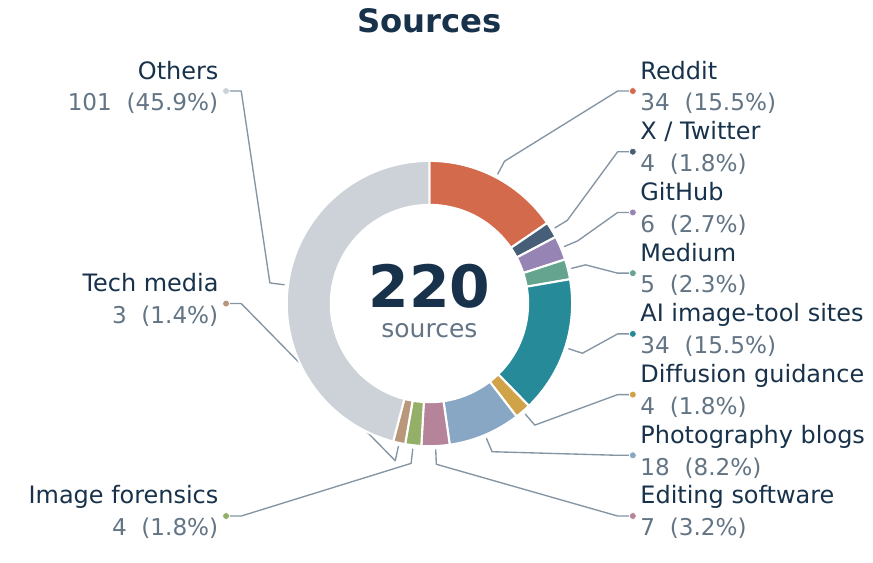}\hfill
\includegraphics[width=0.45\textwidth]{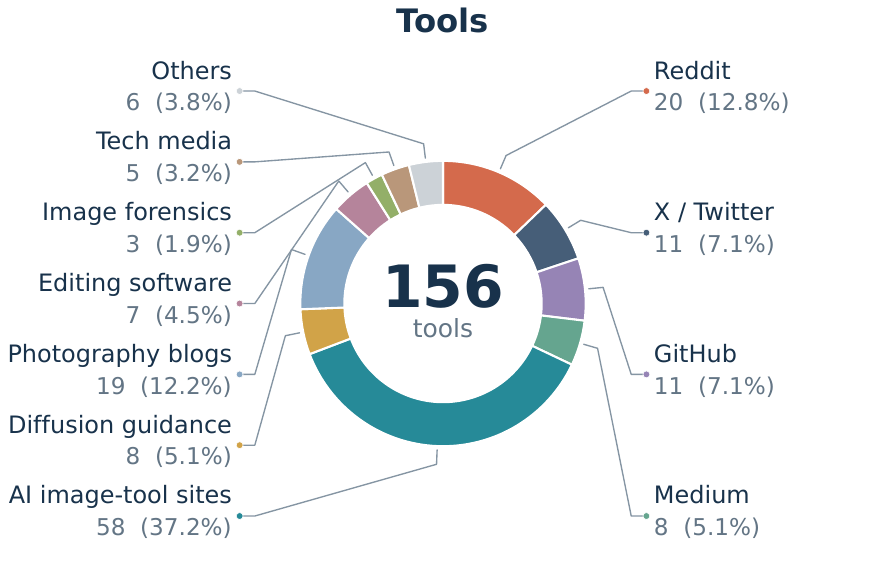}
\caption{{Source and tool composition by source category. Each label reports the count and its percentage of all $220$ collected sources (upper) or $156$ validated tools (lower). Colors identify the same categories in both charts. Percentages are rounded to one decimal place. Full source counts, per-source yields, and largest contributors are retained in Table~\ref{tab:source_categories_full}.}}
\label{fig:source_distributions}
\label{tab:source_categories}
\end{figure}

\begin{table*}[t]
\centering
\caption{{Single-tool ASR on initially detected images. $|\mathcal{B}_d|$ is number of images which are detectable for sepcific detectors among $105$ inputs. Max and Mean summarize the web-derived suite; Base Max is the strongest of six selected baselines; $N_{>\mathrm{base}}$ counts tools exceeding Base Max. Resizing tools are excluded from the \FakeShield{} comparison, following the original evaluation's output-quality exclusion.}}
\label{tab:single_tool_asr}
\begin{tabular}{@{}llrrrrr@{}}
\toprule
& Detector & {$|\mathcal{B}_d|$} & Max ASR & Mean ASR & {Base Max} & {$N_{>\mathrm{base}}$}\\
\midrule
\multirow{4}{*}{VLM}
& \FakeVLM{} & {101} & 45.54\% & 3.22\% & 6.93\% & {16}\\
& \SIDA{} & {99} & 25.25\% & 4.84\% & 12.12\% & {10}\\
& \FakeShield{} & {64} & 46.88\% & 11.60\% & 39.06\% & {3}\\
& \Luna{} & {72} & 66.67\% & 21.87\% & 33.80\% & {14}\\
\midrule
\multirow{2}{*}{Classifier}
& \CommFor{} & {48} & 95.83\% & 21.77\% & 41.67\% & {23}\\
& \AIDE{} & {85} & 100.00\% & 23.10\% & 55.29\% & {25}\\
\bottomrule
\end{tabular}
\end{table*}

\section{Benchmark Results}
In this section, we first conduct some analysis on our obtained WildFakeToolBench and then conduct evaluations on current detectors the collected tools in our benchmark to show the risks of state-of-the-art detectors.
\subsection{Benchmark Source Analysis}
\label{sec:benchmark-sources}
First, we analyze the contribution of different sources in our benchmark collection.
Figure~\ref{fig:source_distributions} distinguishes absolute contribution from tools per collected source. 
From the figure, one can see that, AI image-tool sites supply the largest group in our collection, $58$ tools from $34$ sources. Tools/Src exceed one is because a source may contain several methods.
Besides AI image-tool sites, Reddit also supplies $20$ tools from $34$ related sources, and photography blogs $19$ from $18$. 
We also notice some small groups also generate a lot of tools: X/Twitter contributes $11$ tools from four sources, diffusion-guidance sites eight from four, and GitHub $11$ from six. The possible reason is those sources collected from these websites are more detailed and technical and are easy for implementation, like GitHub.
Besides the famous sites or technical domains, we also collect a lot of related information from other wide websites, with $101$ collected sources. 
However, the knowledge of these domains is not effective for fake-to-real attacks, mainly because there are only general discussions on AI-images with only $6$ accessible tools finally generated.

\begin{table*}[!t]
\centering
\caption{{Full benchmark source statistics. Named platforms are listed separately; other sites are grouped by content type. Tools/Src may exceed one because a source can provide several methods.}}
\label{tab:source_categories_full}
\scalebox{1.0}{%
\begin{tabular}{@{}lrrrl@{}}
\toprule
Source category & Src. & Tools & Tools/Src & Largest contributor \\
\midrule
Reddit & 34 & 20 & 0.59 & \code{reddit.com} (20) \\
X / Twitter & 4 & 11 & 2.75 & \code{x.com} (11) \\
GitHub & 6 & 11 & 1.83 & \code{github.com} (11) \\
Medium & 5 & 8 & 1.60 & \code{medium.com} (7) \\
AI image-tool sites & 34 & 58 & 1.71 & \code{imagera.ai} (10) \\
Diffusion guidance & 4 & 8 & 2.00 & \code{stable-diffusion-art.com} (8) \\
Photography blogs & 18 & 19 & 1.06 & \code{craigboehman.com} (4) \\
Editing software & 7 & 7 & 1.00 & \code{vsco.co} (2) \\
Image forensics & 4 & 3 & 0.75 & \code{ampedsoftware} (2) \\
Tech media & 3 & 5 & 1.67 & \code{kdnuggets.com} (4) \\
Others & 101 & 6 & 0.06 & \code{qwe.edu.pl} (5) \\
\midrule
Total & 220 & 156 & 0.71 & --- \\
\bottomrule
\end{tabular}
}
\end{table*}

\paragraph{Source Contribution.}
We list the sources with largest contributors in Table~\ref{tab:source_categories_full}.
From the results, one can see that the sources of our benchmark include terms from both specialized image-processing resources and general community discussions.
Within AI image-tool sites, \code{imagera.ai} supplies ten tools. 
Other leading contributors are \code{medium.com} with seven, \code{stable-diffusion-art.com} with eight, \code{craigboehman.com} and \code{kdnuggets.com} with four each, and \code{vsco.co} with two. The concentration of sources for our collected attack tools reflects the non-negligible impact of those publicly accessible websites, malicious actors can get attack techniques from those domains.

\paragraph{Tools Production.}
Editing-software sources produce $1.0$ tool per source and image-forensics sources $0.75$. Moreover, Medium and tech media exceed $1.5$. 
Specifically, although AI image-tool sites provide the largest tool pool, some smaller source groups also contain several actionable methods per page. 
Reddit and photography blogs contribute substantial coverage despite lower yields than X/Twitter or diffusion-guidance sites. Besides the tool generation analysis, we also analyze the detector-specific effectiveness separately in Section~\ref{sec_source}.

\begin{table*}[t]
\centering
\caption{Candidate-coverage results for all two-, three-, and four-tool available sets selected from the fixed eight-tool pool defined in Section~\ref{sec_mult}. The evaluation includes $28$, $56$, and $70$ sets of the respective sizes. \emph{Max}: the highest stacked ASR among available tool sets of that size. \emph{$>$25\%}/\emph{$>$50\%}: the number of sets exceeding the corresponding ASR threshold.}
\label{tab:stack_asr}
\scalebox{1.0}{
\begin{tabular}{@{}llccccccccc@{}}
\toprule
&& \multicolumn{3}{c}{Two} & \multicolumn{3}{c}{Three} & \multicolumn{3}{c}{Four} \\
\cmidrule(lr){3-5}\cmidrule(lr){6-8}\cmidrule(lr){9-11}
&Detector & Max & $>$25\% & $>$50\% & Max & $>$25\% & $>$50\% & Max & $>$25\% & $>$50\% \\
\midrule
\multirow{4}{*}{VLM}&\FakeVLM{}    & 63.37\%  & 16 & 6  & 84.16\%  & 46 & 30 & 92.08\%  & 65 & 63 \\
&\SIDA{}       & 49.49\%  & 15  & 0  & 58.59\%  & 55 & 12  & 68.69\%  & 70 & 52  \\
&\FakeShield{} & 64.06\%  & 20 & 4  & 71.88\%  & 55 & 27  & 79.69\%  & 70 & 70  \\
&\Luna{}       & 69.01\%  & 28 & 23  & 81.69\%  & 56 & 56 & 87.32\%  & 70 & 70 \\
\midrule
\multirow{2}{*}{Neural}&\CommFor{}    & 100.00\% & 28 & 27 & 100.00\% & 56 & 56 & 100.00\%  & 70 & 70 \\
&\AIDE{}       & 100.00\% & 25 & 25 & 100.00\% & 56 & 56 & 100.00\% & 70 & 70 \\

\bottomrule
\end{tabular}
}
\end{table*}

\begin{figure}[t]
\centering
\includegraphics[width=\columnwidth]{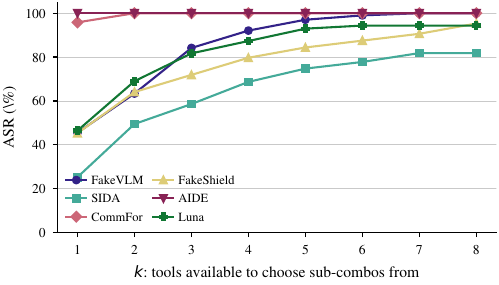}
\caption{{Maximum candidate coverage (stacked ASR) over $k$-tool available sets from the performance-selected eight-tool pool.}}
\label{fig:union_curve}
\end{figure}

\subsection{Single-Tool Effectiveness}
We apply the collected tools in our benchmark respectively on the evaluated $105$ generated images on six state-of-the-art detectors. The summarized results are listed in Table~\ref{tab:single_tool_asr}.

\paragraph{Max Single-Tool ASR Reveal Substantial Vulnerabilities.}
The strongest single-tool ASRs in Table~\ref{tab:single_tool_asr} range from $25.25\%$ against \SIDA{} to $100\%$ against \AIDE{}. 
The highest ASRs are $45.54\%$ for \FakeVLM{}, $46.88\%$ for \FakeShield{}, and $66.67\%$ for \Luna{}.  These results demonstrate that in-the-wild malicious actors could find effective single tools through internet knowledge conversion.

\paragraph{Lot Web-derived Tools Outperforms Academic Baselines.}
Besides the best performed web-derived tools, there also exist a lot effective tools. Compared with the six selected academic baselines, each detectors are vulnerable to a set of web-derived tools whose ASR are higher. For example, over $10$ tools achieves higher ASR than the best academic baselines for five evalauted detectors except \FakeShield{}, but \FakeShield{} also have $3$ tools whose ASR is higher than $40\%$. These results demonstrate that the effective tools are not rarely exist in the internet.

\paragraph{Tool Effectiveness Varies Across Detectors.}
Although our obtained tool bench contain serveral effective tools for detectors, not all of them are effective. Morevoer, tool performance are also different across different detectors. For example, \toolname{print\_rescan} reaches $25.25\%$ against \SIDA{} and $3.96\%$ against \FakeVLM{}.

\subsection{Results on Combining Multiple Available Tools}
\label{sec_mult}
Besides simply adopting single-tools, malicious attackers may collect several useful tools and combine them for effective attack. To explore such risks, we first select eight effective tools as the candidate pool, which are selected the top-2 effective tools based on its single performance on \FakeVLM{},\SIDA{},\FakeShield{}, \CommFor{} and \AIDE{}. And then we select subsets of them, apply the tool set on the image and then evaluate on the detectors.

\paragraph{Stack Performance Across Different Tool Subsets.} 
Figure~\ref{fig:union_curve} shows that larger tool pools provide more opportunities to find combinations that change a detector's decision. For \FakeVLM{}, maximum coverage rises from $63.37\%$ with two available tools to $84.16\%$ with three and $92.08\%$ with four. Four-tool sets reach $68.69\%$ for \SIDA{}, $79.69\%$ for \FakeShield{}, and $87.32\%$ for \Luna{}. \CommFor{} and \AIDE{} both reach $100\%$. With all eight tools available, coverage reaches $100\%$ for \FakeVLM{}, exceeds $95\%$ for \FakeShield{} and \Luna{}, and exceeds $80\%$ for \SIDA{}. These results show that tools derived from public knowledge can provide effective candidates for detector evasion.

Table~\ref{tab:stack_asr} also shows that effective combinations are common in this selected pool. More than half of the three-tool sets exceed $50\%$ ASR for \FakeVLM{}, \Luna{}, \CommFor{}, and \AIDE{}. For \SIDA{}, 12 of the 56 three-tool sets exceed that threshold. Among four-tool sets, at least 52 of 70 exceed $50\%$ ASR for every detector, and all 70 do so for \FakeShield{}, \Luna{}, \CommFor{}, and \AIDE{}. This suggests that attackers could combine publicly available techniques, translated into tools with an LLM, to find effective options against a range of detectors.

\begin{figure}[t]
\centering
\includegraphics[width=\columnwidth]{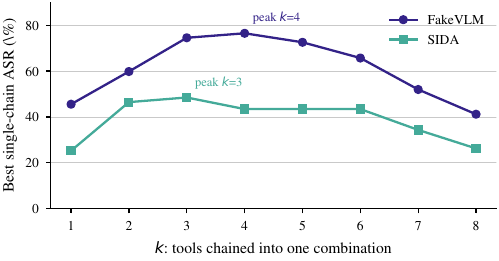}
\caption{{Best reported ASR for applying all $k$ tools sequentially in a chain. Longer chains need not improve performance.}}
\label{fig:stack_no_union}
\end{figure}

\begin{table*}[t]
\centering
\caption{{Operation categories represented in the highest-ASR available sets.} For each detector, \emph{Single} reports the category and ASR of the strongest individual
tool. The two-/three-/four-tool columns report the distinct categories
represented in the highest-ASR available tool set of each size and its
corresponding stacked ASR.
AC=Analog/capture, DS=De-stylization, DR=Detail recovery, CP=Composition,
CW=Color/white balance, RS=Resampling, and LT=Local texture. Categories are
defined in Appendix~\ref{app:categories}.}
\label{tab:category_argmax}
\scalebox{1.0}{
\begin{tabular}{@{}lrrrrrrrr@{}}
\toprule
Detector & Single & ASR & Two & ASR & Three & ASR & Four & ASR \\
\midrule
\FakeVLM{}    & AC & 45.54\% & AC, DS     & 63.37\%  & RS, DS, DR     & 84.16\%  & RS, CP, DS, DR & 92.08\%  \\
\SIDA{}       & AC & 25.25\% & CP, AC     & 49.49\%  & CP, AC, DR     & 58.59\%  & CP, CW, AC, DR & 68.69\%  \\
\FakeShield{} & DS & 46.88\% & CP, CW     & 64.06\%  & CP, AC, CW     & 71.88\%  & CP, AC, CW     & 79.69\%  \\
\Luna{}       & LT & 66.67\% & CP, DR     & 69.01\%  & CP, CW, DR     & 81.69\%  & CP, CW, AC, DR & 87.32\%  \\

\CommFor{}    & CW & 95.83\% & DR, CW     & 100.00\% & DS, DR, CW     & 100.00\% & AC, DS, DR, CW & 100.00\% \\
\AIDE{}       & RS & 100.00\% & RS, DR     & 100.00\% & RS, DS, DR     & 100.00\% & CP, AC, DS, DR & 100.00\% \\

\bottomrule
\end{tabular}
}
\end{table*}
\paragraph{Simultaneously Apply More Tools not Improve Performance.}
Figure~\ref{fig:stack_no_union} instead applies every tool in a chain. From the figure, one can see that the best-chain performance peaks at four tools for \FakeVLM{} and three for \SIDA{}, and the ASR will decrease with the increment of $k$ after that.
The results demonstrate that more tools do not also mean better performance.
The existing validation results also show fewer retained parameter candidates for long chains: approximately four of five pass the SSIM threshold for two-tool processing, compared with approximately two for eight-tool processing. Besides the accumulated distortion, a possible reason for the failure of applying too many tools is that they may affect each other. For example, guassian blurring can mitigate the effect of the photography processing.  

\begin{table*}[!t]
\centering
\caption{{Source platforms represented in the highest-ASR available sets.} For each detector,
\emph{Single} reports the source platform and ASR of the strongest individual
tool. The two-/three-/four-tool columns report the source platforms represented
in the highest-ASR available tool set of each size and its corresponding
stacked ASR.
AI=AI tool sites, RD=Reddit, MD=Medium, ES=Editing software, and TM=Tech
media.}
\label{tab:source_argmax}
\begin{tabular}{@{}lrrrrrrrr@{}}
\toprule
Detector & Single & ASR & Two & ASR & Three & ASR & Four & ASR \\
\midrule
\FakeVLM{}    & AI & 45.54\% & AI, MD     & 63.37\%  & AI, MD, RD & 84.16\%  & AI, RD, MD & 92.08\%  \\
\SIDA{}       & RD & 25.25\% & RD         & 49.49\%  & RD         & 58.59\%  & RD, AI     & 68.69\%  \\
\FakeShield{} & AI & 46.88\% & RD, ES     & 64.06\%  & RD, AI, ES & 71.88\%  & RD, AI, ES & 79.69\%  \\
\CommFor{}    & AI & 95.83\% & RD, ES     & 100.00\% & MD, RD, ES & 100.00\% & RD, MD, ES & 100.00\% \\
\AIDE{}       & AI & 100.00\% & AI, RD     & 100.00\% & AI, MD, RD & 100.00\% & RD, MD     & 100.00\% \\
\Luna{}       & TM & 66.67\% & RD         & 69.01\%  & RD, AI     & 81.69\%  & RD, AI     & 87.32\%  \\
\bottomrule
\end{tabular}
\end{table*}


\subsection{Observed Weaknesses of Current Detectors.}
Our results also point to weaknesses that differ across detectors. Table~\ref{tab:category_argmax} lists the categories of the strongest individual tools and highest-coverage sets.

\paragraph{Single-Aspect Weaknesses} The strongest single tools fall into different categories, suggesting that each detector responds differently to image processing. The strongest tools for \FakeVLM{} and \SIDA{} use analog or capture effects, while those for \FakeShield{}, \Luna{}, \CommFor{}, and \AIDE{} use destylization, local-texture, color/white-balance, and resampling operations, respectively. 

\begin{figure}[t]
\centering
\includegraphics[width=\columnwidth]{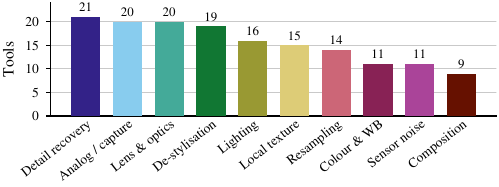}
\caption{{Tool counts across ten assigned processing categories. Categories describe intended operations rather than verified perceptual improvements.}}
\label{fig:category_img}
\end{figure}

\paragraph{Weaknesses from Combination Views.}
Tool combinations also provide new perspectives on detector weaknesses because the strongest individual tool does not always be included in the best combination. For \FakeVLM{}, the strongest three-tool set combines resampling, de-stylization, and detail recovery, reaching $84.16\%$ coverage. Adding composition raises coverage to $92.08\%$ in the strongest four-tool set. However, none of them include the original analog/capture technique.
Although the best combination attack in \SIDA{} includes its related strongest single-tool, results on tool combination more weaknesses of it, especially for the composition attack, which improves the ASR by $25\%$ when including it.
The strongest portfolios for \FakeShield{} and \Luna{} also differ from their best individual tools. \FakeShield{}'s strongest single tool uses de-stylization, but its four-tool set spans composition, color/white balance, and analog/capture. For \Luna{}, local texture leads among single tools, while the strongest four-tool set includes composition, color/white balance, analog/capture, and detail recovery. These further reveals more weaknesses of current detectors. For \CommFor{} and \AIDE{}, the strongest single tools point to color and resampling as useful stress tests, respectively, though other weaknesses may also exist.

Together, these results point to detector-specific weaknesses and suggest directions for future improvements. They also show why web-sourced advice poses a broad risk: discussions span many aspects of image editing and photography, and can therefore yield tools that target different weaknesses across detectors.

\subsection{Source-Level Effectiveness}
\label{sec_source}
We summarize source membership among the strongest individual tools and highest-ASR tool sets in Table~\ref{tab:source_argmax}. 
AI image-tool sites contribute the strongest single tools for four detectors. Reddit provides the strongest single tool for \SIDA{} and appears in the highest-ASR sets for all six detectors. 
Tech media provides the strongest single tool for \Luna{}. 
These results show that these publicly accessible sources can also generate tools effective against different detectors, highlighting the potential reach of web-sourced risks.
In contrast, GitHub and diffusion-guidance sources have relatively high tool generation but do not appear in the reported highest-ASR sets. One possible explanation is that detector designers already consider techniques from these sources as they are more professional.

\subsection{Engagement and Effectiveness.}
Besides the analysis on attack effectiveness and sources, we also explores whether the relationship between the source engagement and the effectiveness of the extracted tools. Since not all sources have the engagement score, we only provide some case study in this section.
In the Reddit case study, \toolname{natural\_color\_correction} comes from a thread with $56$ upvotes and reaches $75.00\%$ ASR against \CommFor{}. 
A tool from a post with zero upvotes and $20$ comments, \toolname{upscale\_grain\_denoise}, reaches $91.67\%$. These examples show that more engagement may be related to the effectiveness of the extracted tools which should pay more attention as a defender.

\subsection{Example of Processed Images}
\label{example-image}
We also show some images processed by an example tool \toolname{print\_rescan} in our WildFakeToolBench with their original images. From the figure, one can see that the processed images are almost same as its origin, with only a rotation and some slight noise to mimic the scanning process. However, such a small variant can convert over $25\%$ detectable fake images from fake to real on \SIDA{}, demonstrating the vulnerable of current detectors.

\begin{figure}[t]
      \centering
      \includegraphics[width=\columnwidth]{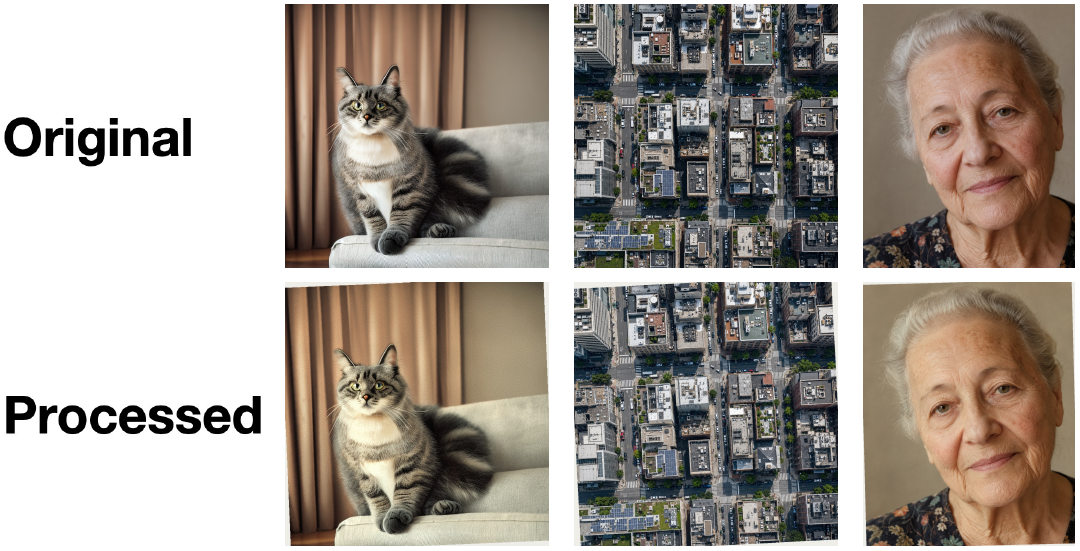}
      \caption{{Original images and the images processed by our collected \toolname{print\_rescan} in the tool bench.}}
\label{fig:example_image}
\end{figure}

\section{Conclusion}
We present \textsc{WildFakeToolBench}, a collection of $156$ executable image-processing tools derived from $220$ public sources across $142$ domains. The tools run on CPUs, retain their source evidence, and are synthesized and validated without detector feedback.

We evaluate the tools on $105$ images from five generators against six detectors. Every detector is vulnerable to at least one tool. The strongest individual tools achieve ASRs from $25.25\%$ to $100\%$ and exceed the strongest of six selected baselines for each detector. An eight-tool pool achieves candidate coverage above $80\%$ for all six detectors. Effectiveness varies by detector, and high coverage does not ensure one-shot evasion. These findings show that web-sourced knowledge can complement researcher-selected robustness evaluations.
\section*{Ethics Considerations}
This work has clear dual-use implications. The transformations collected and
synthesized by \textsc{WildFake} can help researchers identify weaknesses in
AI-image detectors, but they can also be used to make generated images harder
to detect. We therefore present these tools as a robustness-evaluation resource
intended for authorized defensive research and detector development.

Our collection is restricted to content accessible through open-web retrieval,
and we evaluate the resulting tools in a controlled setting on an AI-generated
benchmark. We do not join closed communities, interact with users to obtain
data, or deploy the tools against real-world moderation systems. These measures
reduce, but do not eliminate, the potential risks associated with releasing a
dual-use evaluation resource.

\bibliographystyle{plain}
\bibliography{normal_generated_py3}
\appendix
\section{Construction Details}
\label{app:construction}
\paragraph{Overview.}
\textsc{WildFake} constructs the tool benchmark through retrieval, method extraction and deduplication, code synthesis, and validation. Figure~\ref{fig:benchmark_overview} presents the benchmark overview; the detailed construction procedure is documented below.

\subsection{Retrieval and Corpus Processing}
\label{sec:collect-retrieval}
\label{sec:collect-normalize}
GPT-4o-mini assists with queries covering explicit detector-evasion advice and indirect photography knowledge. Tavily retrieves up to five results per query, including URLs and page text. Platform-specific searches supplement general search for Reddit and X. Collection is limited to public web content; this restriction does not itself determine the licensing of retrieved material. Query strings appear in Appendix~\ref{sec_query}.

We discard pages with fewer than $300$ extracted characters, remove exact-URL duplicates, and suppress domains primarily returning videos, marketplace listings, or other non-text content. Each normalized \code{Post} record stores a stable identifier, page text, URL, domain, matched discourse category, originating query, and available retrieval metadata. Source records support provenance and the source-level analysis.

\subsection{Method Extraction and Deduplication}
\label{sec:collect-mine}
GPT-4o-mini examines sources for actionable processing ideas and their stated mechanisms, retaining quoted evidence. The extraction prompt is reproduced in Appendix~\ref{analyzer-detail}. Methods need not specify exact numeric parameters, since synthesis subsequently supplies concrete implementations.

Deduplication distinguishes repeated concrete techniques from related operations. For example, film-grain overlays and sensor-noise injection need not be merged merely because both add noise. Our deduplication procedure first groups duplicates within batches, then uses compact method summaries for a global pass. Candidate duplicate groups are checked for functional similarity, including comparison of generated implementations. Overlapping verified groups are merged by transitive closure, retaining one representative and source provenance. The duplicate-detection and mechanism-grouping prompts are distinguished in Appendix~\ref{sec_dedup}.

\subsection{Tool Interface and Admission Checks}
\label{sec:distill}
GPT-4o-mini converts each retained idea into a Python class with an image-processing method, a parameter schema, and usage guidance. Implementations use common numeric image-processing libraries and do not load a neural image-editing model. The supplied synthesis prompt prohibits importing PyTorch. Parameters control properties such as noise strength, rotation, and blur; image-specific settings are proposed later during evaluation.

Checks test syntax, the class/interface and parameter dictionary, execution on a fixed validation image, non-identity output, and SSIM-based output distortion. When validation fails, the code and diagnostic are returned to the LLM, with at most three generation attempts per method. The validation image and synthesis prompt are documented in Appendix~\ref{sec_tool_syn}. Validation errors provide software feedback; no detector score or decision is used to repair or admit tools.

\subsection{From Public Sources to Executable Tools}
The collection contains $220$ sources across $142$ domains. Method mining labels $77$ sources valuable and $143$ discard, with no borderline assignments, and extracts $204$ method records. Deduplication removes $35$ redundant records and leaves $169$ records; code synthesis and validation admit $156$ tools, approximately $92\%$ of the deduplicated methods. Of the $13$ failures, five produce identity outputs and eight fail to yield runnable implementations within three attempts. This shows that informal advice can often be converted into executable processing, while also revealing limitations of method extraction and code translation. Conversion success does not imply detector evasion.

\section{{Tool Category Definitions}}
\label{app:categories}

\begin{itemize}
      \item \textbf{Detail Recovery \& Artifact Repair}
      These tools restore local details that image generators often smooth out.
      Typical operations add edge variation, repair blurred regions, and reduce
      the over-smoothed appearance.

      \item \textbf{Analog Medium \& Capture Chain}
      These tools simulate the processing history of physically captured
      images, like adding film grain or in-camera JPEG artifacts. The
      goal is to make a generated image resemble one captured and processed by
      real hardware.

      \item \textbf{Lens and Optical Characteristics}
      These tools simulate imperfections introduced by real lenses, such as 
      chromatic aberration and barrel distortion, targeting the
      unrealistically clean optical appearance of AI images.

      \item \textbf{De-stylisation \& Anti-enhancement}
      These tools reduce common stylized effects of image generators, such as
      beauty-filter smoothing, exaggerated symmetry, and excessive
      enhancement.

      \item \textbf{Lighting Realism}
      These tools improve lighting consistency by adjusting shadows, color
      temperature, and light sources across a scene.

      \item \textbf{Local Texture}
      These tools adjust fine-grained surface textures, such as skin and facial
      details.

      \item \textbf{Resolution \& Resampling}
      These tools simulate the resizing and re-encoding that images processed
      when shared online, such transformations test sensitivity to changes in resolution and interpolation.

      \item \textbf{Color \& White Balance}
      These tools adjust global image properties, including white balance,
      saturation, exposure, and tone mapping, to reproduce the color behavior
      of real photographs.

      \item \textbf{Sensor Noise Injection}
      These tools add noise associated with digital image sensors, like
      PRNU-like patterns, targeting statistical noise associated with digital capture rather than film grain.

      \item \textbf{Composition \& Reframing}
      These tools change framing and composition to mimic casual human
      photography. Examples include off-center subjects, unusual aspect ratios,
      and less deliberate staging.

\end{itemize}

\section{{Detector and Baseline Details}}
\subsection{Detector Configurations}\label{app:detectors}
\begin{description}
  \item[\FakeVLM{}~\cite{WYFKWCWWHL25}] It fine-tunes a LLaVA-1.5-7B on their collected $100$k-image dataset annotated with natural-language artifact clues. Thereby, it can enable the VLM's ability of reasoning to spot the fake features and then recognize the fake images with explanations.
  
  \item[\SIDA{}~\cite{HHLHZPWHC25}] It fine-tunes LISA to emit a classification token and a segmentation token from one forward pass. Then the model can not only classify the fake images, but also localize the patterns of the fake images. Such segmentation ability also helps its classification.

      \item[\FakeShield{}~\cite{XZLTHZ25}] It contains two separately-trained modules: DTE-FDM concatenates a domain-tag classifier's output with image tokens and feeds an LLM for real/fake judgment and explanation; MFLM then takes that explanation plus image tokens to produce a SAM-aligned tamper mask. We only use DTE-FDM for verdict.
      \item[\CommFor{}~\cite{PO25}] It fine-tunes a CLIP-ViT-S backbone to build a binary classifier with $2.7$M images sampled from $4803$ generators.
      \item[\AIDE{}~\cite{YLCHJHX25}] It is a mixture-of-experts detector: DCT-based patch scoring selects the highest- and lowest-frequency patches, whose SRM noise residual is each fed to a dedicated fine-tuned ResNet-50. Moreover, it also combines a separately fine-tuned ConvNeXt-based OpenCLIP branch encodes the whole image or global semantics. Finally, all three pooled features are concatenated into an MLP for the final score.
      \item[\Luna{}~\cite{gpt5_luna}] A hosted vision-language model evaluated by direct prompting. We directly prompt the image and prompt them to classify the image, the prompt is listed in Appendix~\ref{sec_gpt}.
\end{description}
\subsection{Selected Image-Processing Baselines}\label{app:baselines}
\begin{description}

\item[Gaussian or Salt-and-Pepper Noise~\cite{HD19}] This tool applies either additive
Gaussian noise with a randomly drawn standard deviation or salt-and-pepper
impulse noise.

\item[Spectral Suppression~\cite{WRAQ22}] This tool removes periodic spectral peaks using the
FFT and related spectral transformations, based on the observation that such
peaks can serve as fingerprints of generated images.

\item[Defocus Blur~\cite{HD19}] This widely used robustness test applies Gaussian blur with a radius that scales linearly with the strength parameter to mimic optical defocus.

\item[Practical Poisson \& Gaussian Noise~\cite{FTKE08}] This tool applies signal-dependent
Poisson noise together with additive Gaussian noise across the image. It is
designed to model two dominant noise sources of a real camera sensor within one
transformation.

\item[Camera Processing] Motivated by search results on camera processing, we include a commonly used academic method~\cite{BMXCSB19} for this goal. It applies a white-balance scalar, a power-law tone curve for contrast, and Gaussian blur as
a proxy for noise reduction to model a real ISP pipeline.

\item[Reinhard Color Transfer] Motivated by our search results on image
realism, we adopt a widely used academic method~\cite{RAGS01} for this goal. It
matches the colors of the generated image to those of a real reference image.
\end{description}
\section{{Additional Result Details}}\label{app:extended-results}
In this section, we also provide some additional results for our evaluation.

\begin{figure}[h]
\centering
\includegraphics[width=0.9\columnwidth]{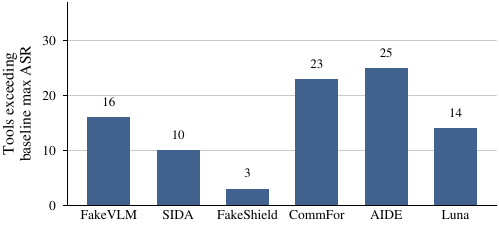}
\caption{Number of released tools whose single-tool ASR exceeds the strongest of
the $6$ academic-sourced baseline tools, per detector.}
\label{fig:baseline_exceed}
\end{figure}

\paragraph{Single-Tool Comparisons.}
Relative to the strongest selected baseline, maximum single-tool ASR improves by $38.61$ percentage points for \FakeVLM{}, $13.13$ for \SIDA{}, $7.82$ for \FakeShield{}, $32.87$ for \Luna{}, $54.16$ for \CommFor{}, and $44.71$ for \AIDE{}. The baseline-exceedance counts are reported in Figure~\ref{fig:baseline_exceed}. Mean single-tool ASRs of $21.77\%$, $23.10\%$, and $21.87\%$ for \CommFor{}, \AIDE{}, and \Luna{}, respectively, indicate that a broader portion of the tested suite affects these systems than affects \FakeVLM{} or \SIDA{} under the current protocol.

\paragraph{Failure-Case Analysis for \SIDA{}.}
We also analyze the $18$ images that remain classified as \textsc{Fake} by
\SIDA{}. These failures are distributed across all $5$ generators: $5$ are
from Midjourney, $4$ each from GPT Image 2 and Qwen-Image, $3$ from Nano Banana
2, and $2$ from Stable Diffusion. No single generator dominates the remaining
failures. Stable Diffusion, despite being the oldest generator in the benchmark,
has the fewest failed images. This distribution suggests that the failures are
not clearly determined by generator source or image quality.
We also compare failures across content categories. The ASRs for document, face
portrait, object, and satellite images are higher than those for animal and
human images. One possible explanation is that \SIDA{} is more robust on animal
and human content.

\paragraph{Example for Source Age and Effectiveness.}
The source for \toolname{print\_rescan} dates to 2023, whereas some methods from 2026 sources reach only approximately $1\%$ ASR. From such examples, we do not find that the relationship effectivness and source age.

\section{{Retrieval Queries and Archived Prompts}}
In this section, we provide queries and prompts for extraction, duplicate detection, mechanism grouping, tool synthesis, parameter selection, and detector evaluation. The grouping prompt is distinct from duplicate detection.

\subsection{Web-Retrieval Queries}
\label{sec_query}
The following queries are used to retrieve explicit detector-evasion advice and
indirect photographic knowledge from the open web.

\begin{tcolorbox}[
  cvprappendixlisting,
  breakable,
  colback=white,
  colframe=gray!60,
  boxrule=0.5pt,
  arc=1pt,
  left=4pt,
  right=4pt,
  top=3pt,
  bottom=3pt,
  title={List of Queries},
  fonttitle=\bfseries
]
\footnotesize
\begin{itemize}
  \item{how to make AI images pass deepfake detector post-processing};
        \item{bypass AI image detector techniques site:photography OR site:reddit};
        \item{AI generated image detection evasion image processing};
        \item{undetectable AI image tutorial 2024 2025}.
  \item{make AI images look less artificial photography tips workflow};
        \item{remove AI sheen look stable diffusion photography};
        \item{make stable diffusion image look like real photograph};
        \item{midjourney photorealistic tips post processing};
        \item{how to make AI art look like real photo tutorial}.
  \item{add film grain analog look AI generated image};
        \item{film emulation analog photography AI image Lightroom};
        \item{Portra 400 film simulation digital photo realistic};
        \item{halation light leak film overlay photography realistic};
        \item{matte lifted blacks fade film look photography}.
  \item{phone camera filter realistic AI image mobile ISP processing};
        \item{iPhone camera noise shadow processing realistic photo};
        \item{mobile camera processing pipeline noise compression artifacts}.
  \item{camera sensor noise PRNU fingerprint photography forensics};
        \item{sensor noise pattern real photograph vs AI image};
        \item{Poisson Gaussian sensor noise model photography}.
  \item{print scan recapture AI image realistic photograph};
        \item{photograph printed AI image realistic laundering};
        \item{recaptured screen image moire pattern forensics}.
  \item{JPEG compression artifacts camera realistic photo AI};
        \item{JPEG quantization table camera identification forensics};
        \item{JPEG ringing blocking artifacts authentic photograph}.
  \item{high frequency texture noise floor real photograph vs AI}.
  \item{frequency separation skin pore texture overlay photography};
        \item{fix plastic AI headshot restore skin pores Photoshop};
        \item{skin texture frequency separation retouching realistic}.
  \item{unsharp mask high-pass filter photography realistic sharpness};
        \item{detail enhancement clarity texture boost photography};
        \item{dithering noise floor photography forensics}.
  \item{chromatic aberration lens fringing photography realistic};
        \item{barrel distortion lens geometry photography realistic};
        \item{vignetting lens corner darkening photography authentic};
        \item{depth of field blur aperture photography realistic AI}.
  \item{white balance temperature tint realistic photograph correction};
        \item{shadow highlight recovery photography tone realistic};
        \item{color grading S-curve split toning cinema photography};
        \item{desaturation HDR tone mapping photography realistic}.
  \item{Poisson blending color harmonization composite photography};
        \item{light harmonization compositing realistic shadows}.
  \item{resample upscale downscale chain image processing photography};
        \item{ComfyUI upscale tile img2img relaundering workflow}.
  \item{GAN fingerprint spectral artifacts detection AI image forensics}.
  \item{adversarial perturbation fool deepfake detector FGSM imperceptible}.
  \item{remove invisible AI watermark SynthID stable signature}.
  \item{EXIF metadata spoofing camera fingerprint injection forensics}.
\end{itemize}
\end{tcolorbox}
\subsection{Method-Mining Prompt}
\label{analyzer-detail}
The analyzer uses the following prompt to identify actionable methods and retain
their supporting evidence from each retrieved source.
\begin{tcblisting}{
  cvprappendixlisting,
  breakable,
  listing only,
  listing engine=listings,
  listing options={
    basicstyle=\ttfamily\footnotesize,
    breaklines=true,
    columns=fullflexible,
    keepspaces=true,
    showstringspaces=false
  },
  colback=white,
  colframe=gray!60,
  boxrule=0.5pt,
  arc=1pt,
  left=4pt,
  right=4pt,
  top=3pt,
  bottom=3pt,
  title={Method-Mining Prompt},
  fonttitle=\bfseries
}
You are mining a community post (Reddit thread or blog/forum article) for IDEAS
that could make AI-generated images pass as real photographs and thereby evade
VLM-based fake-image detectors (FakeVLM, FakeShield, SIDA).

WHAT WE WANT -- the IDEA, not a finished recipe.
We are NOT asking for exact PIL/OpenCV parameters. We want the *concept* and the
*mechanism*: the intuition a practitioner described for why an AI image looks fake
and what they do about it. A downstream coding agent will turn each idea into a
concrete tool later. Example of a good extraction:
  idea: "photograph or re-scan the AI image at a slight tilt, like a phone snapshot
         of a printout"
  mechanism: "introduces perspective skew, re-photography sensor noise and JPEG
              history that real shared photos have but clean AI renders lack"
That is fully acceptable EVEN THOUGH no exact rotation angle is given.

A method may be a single step or many steps -- STEP COUNT DOES NOT MATTER.
Do NOT discard an idea just because it is "only one operation". Judge ideas by
NOVELTY and FEASIBILITY (defined below), not by complexity.

CLASSIFY each extracted idea into one method_type:
  - "generative" : uses a generative model to do the work -- T2I/img2img prompt
                   tricks, negative prompts, low-denoise regeneration, LoRA/
                   checkpoint swaps, diffusion inpainting of artifacts.
  - "numeric"    : a deterministic signal/pixel operation or pipeline -- grain,
                   compression, sensor-noise, color/tone, geometric, capture-pipeline
                   mimicry. No neural model needed at apply time.
  - "hybrid"     : explicitly combines the two (e.g. numeric pipeline THEN a
                   generative regeneration pass, or generate-then-launder).

SKIP only genuinely empty content: venting, news/ethics commentary, pure
"is this fake?" guessing with no technique, or vague aesthetic praise with no
actionable mechanism. Set verdict="discard" and candidate_methods=[] for those.

Score the POST on three axes (1=worst, 10=best):
  - novelty     : how unusual / under-the-radar are these ideas? Favor techniques a
                  detector is UNLIKELY to have been trained against (recent tricks,
                  niche capture pipelines, creative model-side moves). Common,
                  obvious moves score low; clever or fresh ideas score high.
  - feasibility : could an LLM coding agent realistically implement these as an
                  image tool? (a PIL/OpenCV op, or a call to a local generative
                  model). Concrete, mechanism-clear ideas score high; hand-wavy or
                  research-grade ("train a new GAN") ideas score low.
  - relevance   : does the content actually concern making AI images look real /
                  evading detection (vs. unrelated photo editing)?

For each candidate idea capture:
  - name (snake_case, ~2-4 words)
  - idea (one sentence: the concept)
  - mechanism (one sentence: WHY it makes the image look real / why a detector
    might miss it)
  - method_type ("generative" | "numeric" | "hybrid")
  - quoted_evidence (exact phrase you took it from, <= 200 chars)
  - source ("post" | "comment_<author>")

Output ONLY valid JSON, schema:
{
  "post_id": "<echo>",
  "verdict": "valuable" | "borderline" | "discard",
  "scores": {
    "novelty": <int 1-10>,
    "feasibility": <int 1-10>,
    "relevance": <int 1-10>
  },
  "overall_value": <int 1-10>,
  "rationale": "<2-3 sentences>",
  "candidate_methods": [
    {
      "name": "...",
      "idea": "...",
      "mechanism": "...",
      "method_type": "generative" | "numeric" | "hybrid",
      "quoted_evidence": "...",
      "source": "..."
    }
  ]
}

\end{tcblisting}
\subsection{Deduplication Prompts}
\label{sec_dedup}
\paragraph{Batch Deduplication Prompt.}
The following prompt identifies duplicate methods within each processing batch.
\begin{tcblisting}{
  cvprappendixlisting,
  breakable,
  listing only,
  listing engine=listings,
  listing options={
    basicstyle=\ttfamily\footnotesize,
    breaklines=true,
    columns=fullflexible,
    keepspaces=true,
    showstringspaces=false
  },
  colback=white,
  colframe=gray!60,
  boxrule=0.5pt,
  arc=1pt,
  left=4pt,
  right=4pt,
  top=3pt,
  bottom=3pt,
  title={Batch Deduplication Prompt},
  fonttitle=\bfseries
}
You are given a flat list of specific attack-method ideas (each already
individually catalogued), pooled from several independent batches. Because
each batch was processed separately, the SAME concrete technique may appear
more than once under slightly different names/wordings.

Your ONLY job: find groups of ids that describe the SAME concrete technique
(near-identical concept and mechanism -- not just the same broad theme).

BE CONSERVATIVE. Two methods are duplicates only if a downstream coding agent
would end up writing essentially the same tool for both of them. Methods that
are merely related or thematically similar (e.g. "film grain overlay" vs.
"sensor noise injection" vs. "print re-scan blur") are NOT duplicates -- leave
them ungrouped. If you are not confident two ids are the same technique, do
not group them.

Do not invent new ids. Do not include singleton groups (a group only makes
sense with >= 2 members). Most ids will not appear in any group -- that is
expected and correct.

Output ONLY valid JSON, schema:
{
  "duplicate_groups": [
    ["m3", "m17", "m42"],
    ["m8", "m55"]
  ]
}

\end{tcblisting}

\paragraph{{Mechanism-Grouping Prompt.}}
The following archived prompt groups deduplicated methods into higher-level mechanism categories.
\begin{tcblisting}{
  cvprappendixlisting,
  breakable,
  listing only,
  listing engine=listings,
  listing options={
    basicstyle=\ttfamily\footnotesize,
    breaklines=true,
    columns=fullflexible,
    keepspaces=true,
    showstringspaces=false
  },
  colback=white,
  colframe=gray!60,
  boxrule=0.5pt,
  arc=1pt,
  left=4pt,
  right=4pt,
  top=3pt,
  bottom=3pt,
  title={Mechanism-Grouping Prompt},
  fonttitle=\bfseries
}
You are given the FINAL deduplicated list of specific attack-method ideas
(for making AI-generated images pass as real / evade VLM fake-image
detectors). Each has already been individually catalogued and scored
elsewhere -- you are not re-scoring or dropping any of them here.

Your job: group these specific methods into a smaller set of GENERALIZED /
ARCHETYPE methods -- higher-level strategies that explain WHY a cluster of
specific methods works (the shared mechanism), for use in reporting and
strategy selection. This is an ADDITIONAL summary layer on top of the
specific list, not a replacement for it -- every specific method stays in the
specific list regardless of which cluster it lands in here.

Guidelines:
  - Produce however many clusters naturally fit the data -- typically
    somewhere in the 6-15 range for ~100-150 specific methods, fewer if the
    pool is smaller. Do not force a fixed count.
  - Every specific method should belong to exactly one cluster. A method that
    is genuinely singular (no others share its mechanism) may form a
    cluster of one -- that's fine, don't force it into a bad fit.
  - Name each cluster for the shared MECHANISM (e.g. "capture_pipeline_mimicry",
    "signal_domain_texture_injection", "semantic_prompt_evasion"), not for a
    single member's name.
  - `member_names` must reference the exact `name` field of specific methods
    from the input list.
  - `priority` for a cluster = the max priority among its members (rounded to
    1 decimal).

Output ONLY valid JSON, schema:
{
  "generalized_methods": [
    {
      "name": "snake_case_cluster_name",
      "description": "<1-2 sentences: the shared mechanism/strategy>",
      "member_names": ["specific_method_a", "specific_method_b"],
      "priority": <float, 1 decimal>
    }
  ]
}


\end{tcblisting}

\subsection{Tool Synthesis and Validation}
\label{sec_tool_syn}
\paragraph{Tool-Synthesis Prompt.}
The following prompt converts each retained method into a runnable
image-processing tool that follows the framework's common interface.
\begin{tcblisting}{
  cvprappendixlisting,
  breakable,
  listing only,
  listing engine=listings,
  listing options={
    basicstyle=\ttfamily\footnotesize,
    breaklines=true,
    columns=fullflexible,
    keepspaces=true,
    showstringspaces=false
  },
  colback=white,
  colframe=gray!60,
  boxrule=0.5pt,
  arc=1pt,
  left=4pt,
  right=4pt,
  top=3pt,
  bottom=3pt,
  title={Prompt for Tool Synthesis},
  fonttitle=\bfseries
}
You are a Python engineer implementing an image-manipulation tool for an
adversarial-attack research framework. You are given a single ADVERSARIAL IDEA
mined from a community blog/forum (concept + mechanism), and you must turn it
into a concrete, runnable tool.

The tool defeats a fake-image detector by post-processing an AI-generated image
so it reads as a real photograph. Implement the IDEA faithfully -- choose sensible
concrete operations and parameter ranges yourself; the idea gives you the concept,
you supply the realization.

You MUST output a single, self-contained Python module that defines ONE class
subclassing `BaseTool`. Follow this exact contract:

```python
from typing import Dict, Any
from PIL import Image
import numpy as np
# you may also import: scipy.ndimage, scipy.signal, math, random  (all available)
# DO NOT import torch -- numpy + PIL + scipy .
from .base_tool import BaseTool


class <CamelCaseName>Tool(BaseTool):
    def __init__(self):
        super().__init__(
            name="<snake_case_name>",                 # MUST match the given name
            description="<what the transform does, effect-only, 1-3 sentences>",
        )

    def apply(self, image: Image.Image, params: Dict[str, Any]) -> Image.Image:
        self.validate_params(params)
        # read params with .get(name, default); convert image to RGB if needed;
        # operate via numpy/PIL/scipy/cv2; return a NEW PIL.Image (mode "RGB").
        ...

    def get_param_schema(self) -> Dict[str, Any]:
        # 1-4 params, each: {"type": "int|float|str", "min":.., "max":..,
        #                    "description":.., "default":.., "required": False}
        ...

    def get_guidance(self) -> str:
        return "<one line: parameter ranges + when this tool is most useful>"

```

\end{tcblisting}
\paragraph{Validation Image and Retry Procedure.}
The test image is not iid noise. A $24\times24$ uniform random field is
bicubically upsampled to $512\times512$ and a $\sigma=8$ Gaussian grain is added
on top, under a fixed \code{RandomState(42)}. The reason is on pure iid noise there is no spatial correlation for a blur,
geometric or texture-matching operation to preserve, so SSIM measurement cannot be done. If validation
fails, the generated tool and the failure reason are returned to the LLM for
revision. We allow at most $3$ generation attempts per method.

\subsection{Tool-Parameter Selection Prompt}
The following prompt asks GPT-4o-mini to select five image-specific parameter
settings that span the applicable strength range of a tool.
\begin{tcblisting}{
  cvprappendixlisting,
  breakable,
  listing only,
  listing engine=listings,
  listing options={
    basicstyle=\ttfamily\footnotesize,
    breaklines=true,
    columns=fullflexible,
    keepspaces=true,
    showstringspaces=false
  },
  colback=white,
  colframe=gray!60,
  boxrule=0.5pt,
  arc=1pt,
  left=4pt,
  right=4pt,
  top=3pt,
  bottom=3pt,
  title={Tool-Parameter Selection Prompt},
  fonttitle=\bfseries
}
You are proposing parameter settings for an image-editing tool, to test how a fake-image detector responds to different intensities of the SAME tool applied to the SAME image.

Tool: {tool.name}
Description: {tool.description}
Parameters: {_param_desc(tool)}
Usage guidance: {guidance}

You will be shown ONE image. Look at it and propose exactly {N_CANDIDATES} DIVERGING candidate parameter settings for this tool, spanning a clear intensity range from very light to very strong. Each candidate must be genuinely different (not near-duplicate values) so together they explore the tool's effective range on this specific image. Base your choices on what you can see in the image (e.g. its existing quality, texture, lighting) as well as the tool's guidance above.

IMPORTANT: even "very_strong" must stay a PLAUSIBLE edit -- the image must still look like a real, undamaged photo afterward. Do not propose settings so extreme that they visibly destroy the image (heavy artifacting, unrecognizable content, obvious tampering). A candidate that wrecks image quality will be discarded during evaluation regardless of whether it fools the detector, so there is no benefit to being overly aggressive -- stay within the range where the edit could plausibly pass as an unedited photo to a human viewer.

Respond with ONLY valid JSON, no other text:
{"candidates": [
  {"tier": "very_light", "params": {...}, "reasoning": "..."},
  {"tier": "light", "params": {...}, "reasoning": "..."},
  {"tier": "moderate", "params": {...}, "reasoning": "..."},
  {"tier": "strong", "params": {...}, "reasoning": "..."},
  {"tier": "very_strong", "params": {...}, "reasoning": "..."}
]}
\end{tcblisting}

\subsection{GPT-Based Detector Prompt}
\label{sec_gpt}
The following prompt is used to obtain a binary real/fake judgment and a
corresponding probability from \Luna{}.
\begin{tcblisting}{
  cvprappendixlisting,
  breakable,
  listing only,
  listing engine=listings,
  listing options={
    basicstyle=\ttfamily\footnotesize,
    breaklines=true,
    columns=fullflexible,
    keepspaces=true,
    showstringspaces=false
  },
  colback=white,
  colframe=gray!60,
  boxrule=0.5pt,
  arc=1pt,
  left=4pt,
  right=4pt,
  top=3pt,
  bottom=3pt,
  title={GPT-Based AI-Image Detection Prompt},
  fonttitle=\bfseries
}
Estimate the probability that this image is AI-generated or manipulated by AI generators.

This image comes from a mixed pool of real photographs and generated images. You are NOT told the proportion, so do not assume it: judge this image on its own evidence.

Both mistakes cost the same. Calling a real photograph FAKE is as damaging as letting a generated one pass, so do not shade your answer toward either. Use the full 0-100 range and mean it: 90+ only with a clear artifact you can name, 50 for genuinely ambiguous, 10 or below when the image carries positive signs of real capture. If you would be unwilling to defend the number to someone holding the ground truth, move it toward 50.

Reply with ONLY this JSON: {"p_fake": <integer 0-100>, "verdict": "FAKE" or "REAL", "why": "<8 words max>"}. p_fake is your probability the image is AI-generated or manipulated by AI generators; verdict must agree with it (>=50 means FAKE).




\end{tcblisting}

\end{document}